%% file: main.tex
\documentclass[letterpaper, 10 pt, conference]{ieeeconf}
\IEEEoverridecommandlockouts   % to use \thanks
\usepackage{times}
\usepackage{multicol}
\usepackage[bookmarks=true]{hyperref}

\usepackage{datetime2}
\usepackage{latexsym}
\usepackage{amsmath}
\usepackage{amssymb}
\usepackage{array}
\usepackage{multirow}
\usepackage{url}
\usepackage[usenames,dvipsnames]{xcolor}
\usepackage{graphicx}
\usepackage{wrapfig}
\usepackage{tikz}
\usepackage{tikz-qtree}
\let\labelindent\relax
\usepackage{enumitem}
\usepackage{adjustbox}
\usepackage{calc}
\usepackage{caption}
\usepackage{subcaption}
\usepackage[normalem]{ulem}
\usepackage{makecell}
\usepackage{scalefnt}
\usepackage{algorithm2e}
\usepackage{algorithmicx}
\usepackage[noend]{algpseudocode}
\usepackage[makeroom]{cancel}
\usepackage{xspace}
\usepackage{epsfig}
\usepackage{xcolor}
\usepackage{booktabs}
\usepackage{flushend}
\usepackage{cite}

\usepackage[colorinlistoftodos,prependcaption,textsize=tiny]{todonotes}

\begin{document}

\title{Improving Grounded Natural Language Understanding\\through Human-Robot Dialog}

\author{Jesse Thomason$^*$, Aishwarya Padmakumar$^\dagger$, Jivko Sinapov$^\ddagger$, Nick Walker$^*$, Yuqian Jiang$^\dagger$,\\Harel Yedidsion$^\dagger$, Justin Hart$^\dagger$, Peter Stone$^\dagger$, and Raymond J. Mooney$^\dagger$
\thanks{$^{*}$Paul G. Allen School of Computer Science and Engineering, University of Washington. \texttt{jdtho@cs.washington.edu}}
\thanks{$^{\dagger}$ Department of Computer Science, University of Texas at Austin.}
\thanks{$^{\ddagger}$Department of Computer Science, Tufts University.}
}

\maketitle

\begin{abstract}
Natural language understanding for robotics can require substantial domain- and platform-specific engineering.
For example, for mobile robots to pick-and-place objects in an environment to satisfy human commands, we can specify the language humans use to issue such commands, and connect concept words like \emph{red can} to physical object properties.
One way to alleviate this engineering for a new domain is to enable robots in human environments to adapt dynamically---continually learning new language constructions and perceptual concepts.
In this work, we present an end-to-end pipeline for translating natural language commands to discrete robot actions, and use clarification dialogs to jointly improve language parsing and concept grounding.
We train and evaluate this agent in a virtual setting on Amazon Mechanical Turk, and we transfer the learned agent to a physical robot platform to demonstrate it in the real world.
\end{abstract}

\section{Introduction}
\label{sec:introduction}
\input{01_introduction.tex}

\section{Related Work}
\label{sec:related_work}
\input{02_related_work.tex}

\section{Conversational Agent}
\label{sec:agent}
\input{03_agent.tex}

\section{Experiments}
\label{sec:experiments}
\input{04_experiments.tex}

\section{Conclusion}
\label{sec:conclusion}
\input{05_conclusion.tex}

\section*{Acknowledgements}
This work was supported by a National Science Foundation Graduate Research Fellowship to the first author, an NSF EAGER grant (IIS-1548567), and an NSF NRI grant (IIS-1637736).
This work has taken place in the Learning Agents Research Group (LARG) at UT Austin.
LARG research is supported in part by NSF (CNS-1305287, IIS-1637736, IIS-1651089, IIS-1724157), TxDOT, Intel, Raytheon, and Lockheed Martin.

\bibliographystyle{IEEEtran}
\bibliography{main}

\end{document}

%% file: 01_introduction.tex
% Motivation - robots
As robots become ubiquitous across diverse human environments such as homes, factory floors, and hospitals, the need for effective human-robot communication grows.
These spaces involve domain-specific words and affordances, e.g., \emph{turn on the kitchen lights}, \emph{move the pallet six feet to the north}, and \emph{notify me if the patient's condition changes}.
Thus, pre-programming robots' language understanding can require costly domain- and platform-specific engineering.
In this paper, we propose and evaluate a robot agent that leverages conversations with humans to expand an initially low-resource, hand-crafted language understanding pipeline to reach better common ground with its human partners.

% Language grounding
% Some world knowledge needs to be acquired on the fly, such as whether the new word \emph{rattling} applies to nearby objects.
% Gathering correspondences between physical objects and perceptual concepts (e.g., \emph{heavy, brown, mug}) is a time-consuming annotation effort if performed exhaustively by humans.
% We instead extract these correspondences from human-robot conversations as needed (Figure~\ref{fig:rattling_highlight}) using an active learning strategy that builds on prior work~\cite{thomason:corl17}.
% Complementing this, we show how the agent can improve its semantic parser using weak supervision from human-robot conversations to acquire new grammatical constructions (extending prior work~\cite{thomason:ijcai15}) and examples of new concept word usage.

We combine bootstrapping better semantic parsing through signal from clarification dialogs~\cite{thomason:ijcai15}, previously using no sensory representation of objects, with an active learning approach for acquiring such concepts~\cite{thomason:corl17}, previously restricted to object identification tasks.
Thus, our system is able to execute natural language commands like {\it Move a rattling container from the lounge by the conference room to Bob's office} (Figure~\ref{fig:segbot_demo}) that contain compositional language (e.g., {\it lounge by the conference room} understood by the semantic parser and objects to be identified by their physical properties (e.g., {\it rattling container}).
The system is initialized with a small seed of natural language data for semantic parsing, and no initial labels tying concept words to physical objects, instead learning parsing and grounding as needed through human-robot dialog.

\begin{figure}
\centering
  \includegraphics[width=0.49\linewidth]{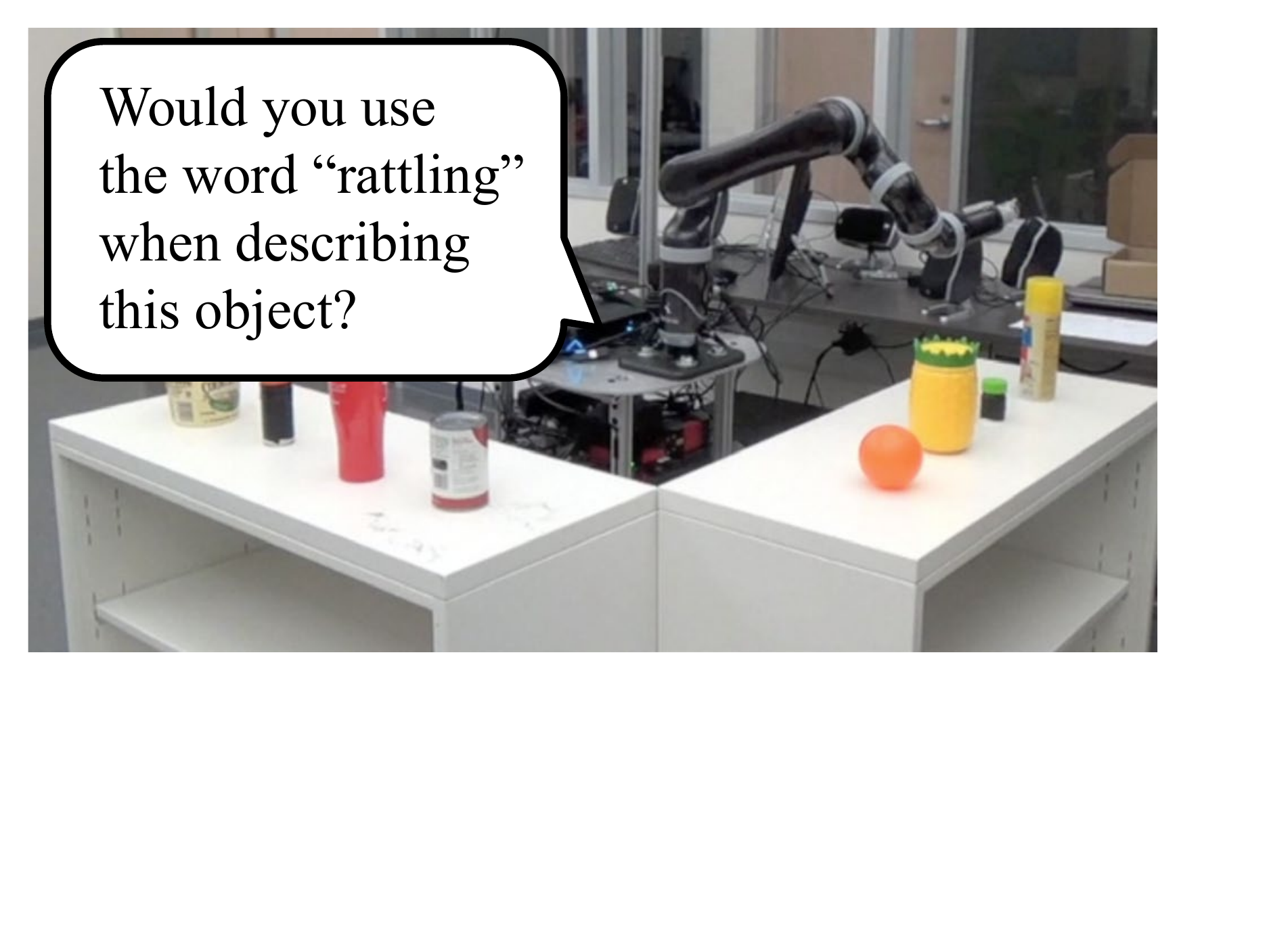}
  \includegraphics[width=0.49\linewidth]{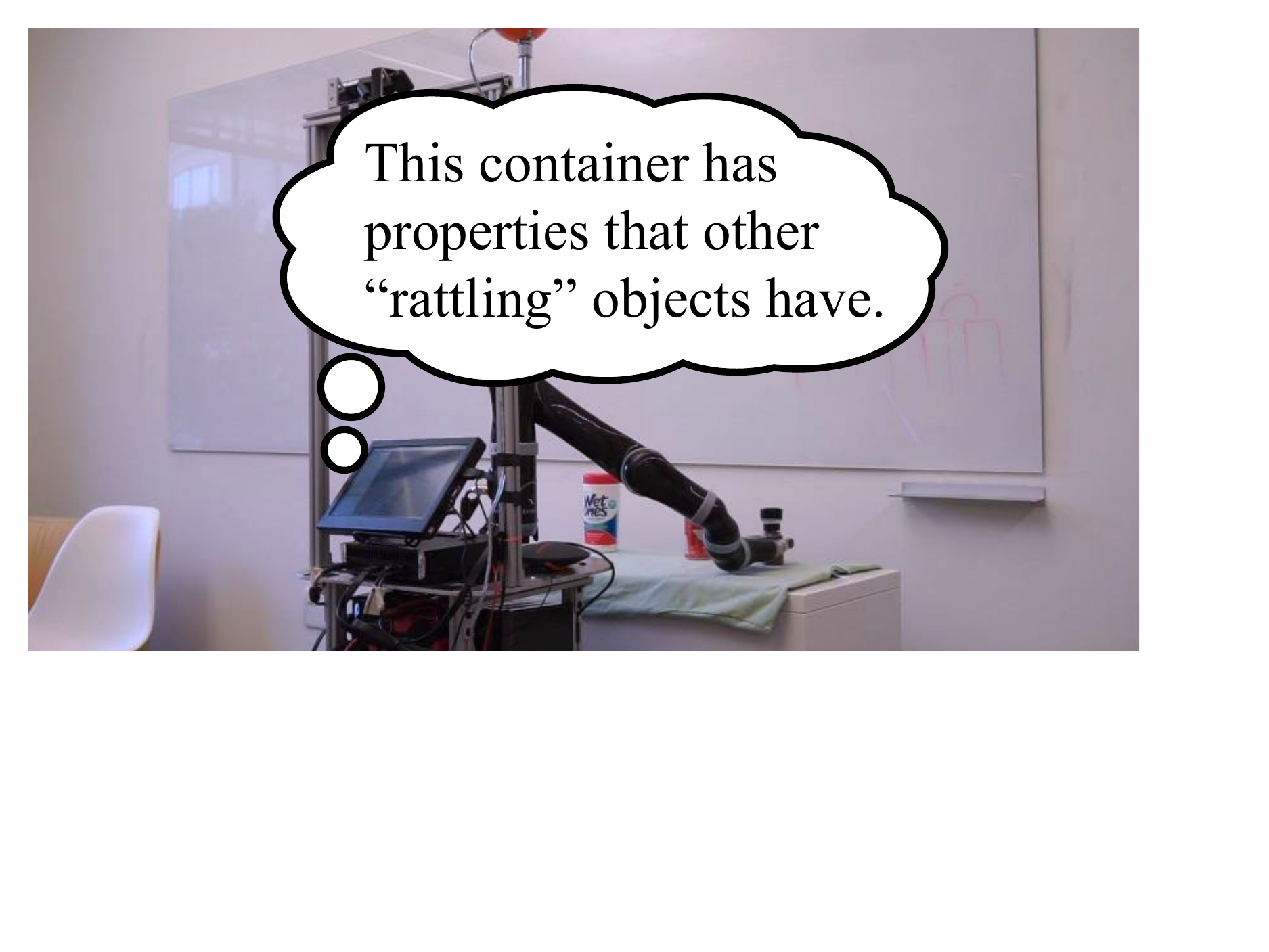}
  \vspace{-1.5cm}
\caption{Through dialog, a robot agent can acquire task-relevant information from a human on the fly.
Here, \emph{rattling} is a new concept the agent learns with human guidance in order to pick out a remote target object later on.
}
  \label{fig:rattling_highlight}
\end{figure}

Our contributions are: 1) a dialog strategy to improve language understanding given only a small amount of initial in-domain training data; 2) dialog questions to acquire perceptual concepts \emph{in situ} rather than from pre-labeled data or past interactions alone (Figure~\ref{fig:rattling_highlight}); and 3) a deployment of our dialog agent on a full stack, physical robot platform.

% Experiments
We evaluate this agent's learning capabilities and usability on Mechanical Turk, asking human users to instruct the agent through dialog to perform three tasks: navigation (\emph{Go to the lounge by the kitchen}), delivery (\emph{Bring a red can to Bob}), and relocation (\emph{Move an empty jar from the lounge by the kitchen to Alice's office}).
We find that the agent receives higher qualitative ratings after training on information extracted from previous conversations.
We then transfer the trained agent to a physical robot to demonstrate its continual learning process in a live human-robot dialog.\footnote{A demonstration video can be viewed at\\ \url{https://youtu.be/PbOfteZ_CJc?t=5}.}

\begin{figure*}[ht]
\centering
  \includegraphics[width=0.9\linewidth]{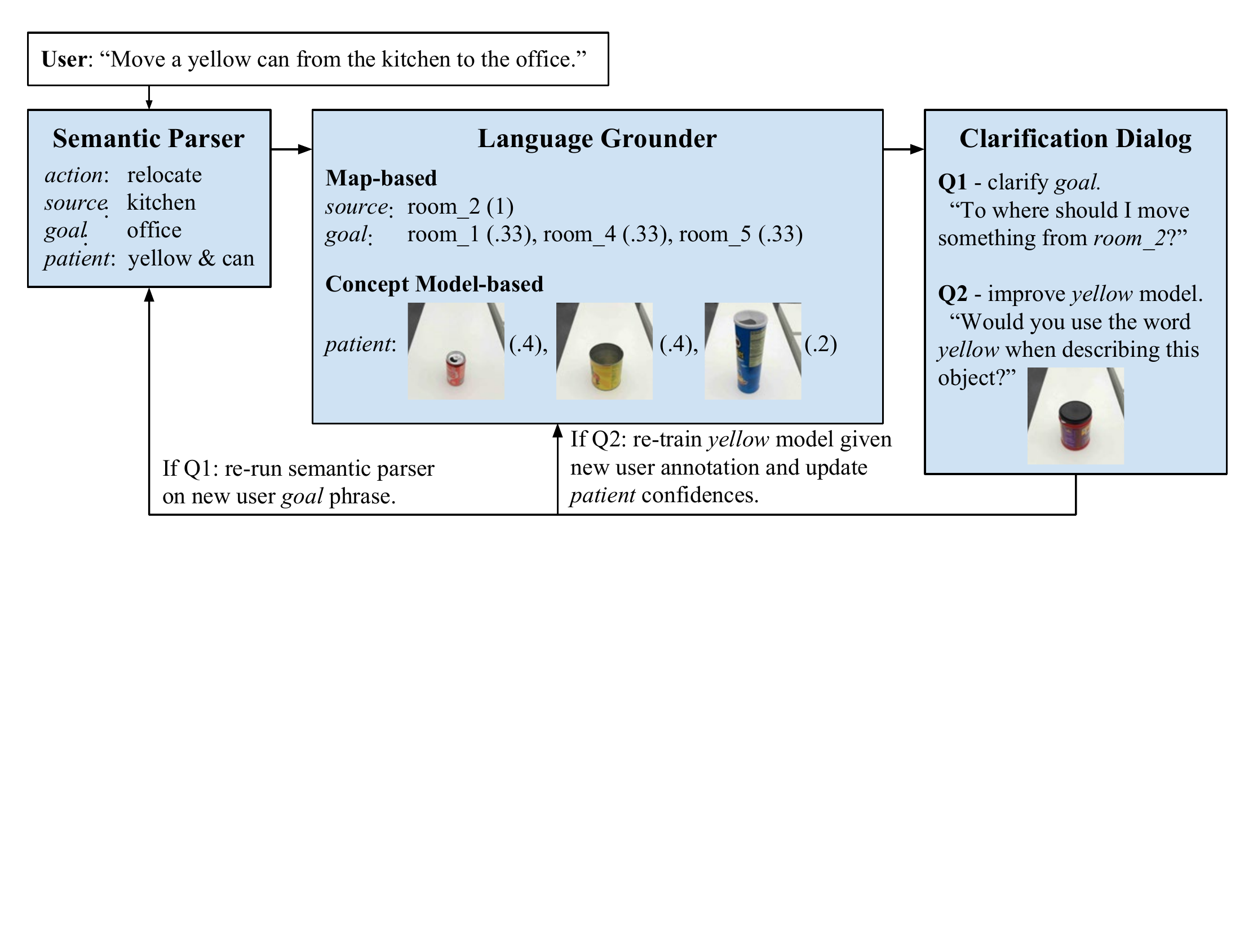}
  \vspace{-5cm}
\caption{User commands are parsed into semantic slots ({\bf left}), which are grounded ({\bf center}) using either a known map (for rooms and people) or learned concept models (for objects) to a distribution over possible satisfying constants (e.g., all rooms that can be described as an ``office'').
A clarification dialog ({\bf right}) is used to recover from ambiguous or misunderstood slots (e.g., {\it Q1}), and to improve concept models on the fly (e.g., {\it Q2}).}
  \label{fig:phm_components}
\end{figure*}

%% file: 02_related_work.tex
% Motivate commanding.
% Perception and action with semantic parsing.
Research on the topic of humans instructing robots spans natural language understanding, vision, and robotics.
Recent methods perform semantic parsing using sequence-to-sequence~\cite{kovcisky:emnlp16,jia:acl16,konstas:acl17} or sequence-to-tree~\cite{dong:acl16} neural networks, but these require hundreds to thousands of examples.
In human-robot dialog, gathering information at scale for a given environment and platform is unrealistic, since each data point comes from a human user having a dialog interaction in the same space as a robot.
Thus, our methods assume only a small amount of seed data.

Semantic parsing has been used as a language understanding step in tasks involving unconstrained natural language instruction, where a robot must navigate an unseen environment~\cite{kollar:hri10,matuszek:iser12,anderson:cvpr18,wang:eccv18,shah:icra18}, to generate language requests regarding a shared environment~\cite{tellex:rss14}, and to tie language to planning~\cite{williams:icra18,skoviera:iros18,chai:ijcai18,nyga:corl18}.
Other work memorizes new semantic referents in a dialog, like \emph{this is my snack}~\cite{paul:ijcai17}, but does not learn a general concept for {\it snack}.
In this work, our agent can learn new referring expressions and novel perceptual concepts on the fly through dialog.

% Motivate and define grounding.
% Interactive grounding through human-robot dialog.
Mapping from a referring expression such as \emph{the red cup} to a referent in the world is an example of the \textit{symbol grounding problem}~\cite{harnad:phys90}.
\textit{Grounded language learning} bridges machine representations with natural language.
Most work in this space has relied solely on visual perception ~\cite{matuszek:icml12,fitzgerald:emnlp13,krishnamurthy:acl13,zitnik:cvpr13,bisk:aaai18,pillai:aaai18}, though some work explores grounding using audio, haptic, and tactile signals produced when interacting with an object~\cite{chu:icra13,orhan2013co,nakamura2014mutual,gao:icra16}.
In this work, we explicitly model perceptual predicates that refer to visual (e.g., \emph{red}), audio (e.g., \emph{rattling}), and haptic properties (e.g., \emph{full}) of a fixed set of objects.
We gather data for this kind of perceptual grounding using interaction with humans, following previous work on learning to ground object attributes and names through dialog~\cite{vogel:aaai10,parde:ijcai15,thomason:ijcai16,yang:corl18,vanzo:aamas18}.
We take the additional step of using these concepts to accomplish a room-to-room, pick-and-place task instructed via a human-robot dialog.
To our knowledge, there is no existing end-to-end, grounded dialog agent with multi-modal perception against which to compare, and we instead ablate our model during evaluation.

%% file: 03_agent.tex
% Overview.
We present a end-to-end pipeline (Figure~\ref{fig:phm_components}) for an task-driven dialog agent that fulfills requests in natural language.\footnote{The source code for this dialog agent, as well as the deployments described in the following section, can be found at\\ \url{https://github.com/thomason-jesse/grounded_dialog_agent}.}
% The agent uses semantic parsing as an understanding step and learns multi-modal perceptual concept models to connect language to physical objects.

\subsection{Semantic Parser}
\label{ssec:agent_semantic_parser}
\input{0301_agent_semantic_parser.tex}

\begin{table*}[ht]
\centering
\begin{tabular}[ht!]{|ll|ll|}
	\hline
	\bf $B$ max per role & \bf Min & & \\
	\bf (\emph{action}, \emph{patient}, & \bf Prob & \bf Question & \bf Type \\
	\bf \emph{recipient}, \emph{source}, \emph{goal}) & \bf $B$ Role & & \\ \hline
	$(\varnothing,\varnothing,\varnothing,\varnothing,\varnothing)$ & All & What should I do? & Clarification \\
	$(\text{walk},\varnothing,\varnothing,\varnothing,r_1)$ & \emph{action} & You want me to go somewhere? & Confirmation \\
	$(\text{deliver},\varnothing,p_1,\varnothing,\varnothing)$ & \emph{patient} & What should I deliver to $p_1$? & Clarification \\
	$(\text{relocate}, \varnothing,\varnothing,\varnothing,\varnothing)$ & \emph{source} & Where should I move something from on its way somewhere else? & Clarification \\
	$(\text{relocate}, o_1,\varnothing,r_1,r_2)$ & - & You want me to move $o_1$ from $r_1$ to $r_2$? & Confirmation \\
	\hline
\end{tabular}
\caption{Samples of the agent's static dialog policy $\pi$ for mapping belief states to questions.}
\label{tab:phm_dialog_policy_examples}
\end{table*}

\subsection{Language Grounding}
\label{ssec:agent_grounding}
\input{0302_agent_grounding.tex}

\subsection{Clarification Dialog}
\label{ssec:agent_dialog}
\input{0303_agent_dialog.tex}

\subsection{Learning from Conversations}
\label{ssec:agent_learning}
\input{0304_agent_learning.tex}

%% file: 0301_agent_semantic_parser.tex
The semantic parsing component takes in a sequence of words and infers a semantic meaning representation of the task.
For example, a {\it relocate} task moves an item ({\it patient}) from one place ({\it source}) to another ({\it goal}) (Figure~\ref{fig:phm_components}).
The agent uses the Combinatory Categorial Grammar (CCG) formalism~\cite{steedman:nts11} to facilitate parsing.

% , and uncovers new lexical entries during training from past dialogs (Section~\ref{ssec:agent_learning}).
% The agent discovers new perceptual concepts on the fly during conversations with human users (e.g., when the word \emph{red} is first seen).

Word embeddings~\cite{mikolov:nips13} augment the lexicon at test time to recover from out-of-vocabulary words, an idea similar in spirit to previous work~\cite{bastianelli:ijcai16}, but taken a step further via formal integration into the agent's parsing pipeline.
This allows, for example, the agent to use the meaning of known word \emph{take} for unseen word \emph{grab} at inference time.

%% file: 0302_agent_grounding.tex
The grounding component takes in a semantic meaning representation and infers denotations and associated confidence values (Figure~\ref{fig:phm_components}).
The same semantic meaning can \emph{ground} differently depending on the environment.
For example, \emph{the office by the kitchen} refers to a physical location, but that location depends on the building.

Perceptual concepts like \emph{red} and \emph{heavy} require considering sensory perception of physical objects.
The agent builds multi-modal feature representations of objects by exploring them with a fixed set of behaviors.
In particular, before our experiments, a robot performed a \textit{grasp}, \textit{lift}, \textit{lower}, \textit{drop}, \textit{press}, and \textit{push} behavior on every object, recording \textit{audio} information from an onboard microphone and \textit{haptic} information from force sensors in its arm.
That robot also \textit{look}ed at each object with an RGB camera to get a visual representation.
Summary audio, haptic, and visual features are created for each applicable behavior (e.g., \textit{drop-audio}, \textit{look-vision}), and these features represent objects at training and inference time both in simulation and the real world.\footnote{That is, at inference time, while all objects have been explored, the language concepts that apply to them (e.g., {\it heavy}) must be inferred from their feature representations.}

Feature representations of objects are connected to language labels by learning discriminative classifiers for each concept using the methods described in previous work~\cite{sinapov:icra14,thomason:ijcai16}.
In short, each concept is represented as an ensemble of classifiers over behavior-modality spaces weighted according to accuracy on available data (so {\it yellow} weighs \textit{look-vision} highly, while {\it rattle} weighs \textit{drop-audio} highly).
While the objects have already been explored (i.e., they have feature representations), language labels must be gathered on the fly from human users to connect these features to words.

Different predicates afford the agent different certainties.
Map-based facts such as room types (\emph{office}) can be grounded with full confidence.
For words like \emph{red}, perceptual concept models give both a decision and a confidence value in~$[0, 1]$.
Since there are multiple possible groundings for ambiguous utterances like \emph{the office}, and varied confidences for perceptual concept models on different objects, we associate a confidence distribution with the possible groundings for a semantic parse (Figure~\ref{fig:phm_components}).

%% file: 0303_agent_dialog.tex
% \begin{figure*}[ht]
% \centering
%   \includegraphics[width=1.\linewidth]{figures/belief_update_recipient.png}
% \caption{Example belief from the question \emph{To whom should I bring something?} being answered \emph{alice}.}
%   \label{fig:phm_belief_update_example}
% \end{figure*}

% Clarification policy similar to that used in ijcai15
We denote a dialog agent with $\mathcal{A}$.
Dialog begins with a human user commanding the agent to perform a task, e.g., \emph{grab the silver can for alice}.
The agent maintains a belief state modeling the unobserved true task in the user's mind, and uses the language signals from the user to infer that task.
The command is processed by the semantic parsing and grounding components to obtain pairs of denotations and their confidence values.
Using these pairs, the agent's belief state is updated, and it engages in a clarification dialog to refine that belief (Figure~\ref{fig:phm_components}).

% Belief state
The belief state, $\mathcal{B}$, is a mapping from semantic roles (components of the task) to probability distributions over the known constants that can fill those roles (\emph{action}, \emph{patient}, \emph{recipient}, \emph{source}, and \emph{goal}).
The belief state models uncertainties from both the semantic parsing (e.g., prepositional ambiguity in ``pod by the office to the north''; is the pod or the office north?) and language grounding (e.g., noisy concept models) steps of language understanding.

% role is initialized to uniform probability across three robot actions (\emph{walk}, \emph{deliver}, and \emph{relocate}).
% The remaining roles' belief states are initialized with half of their probability mass on an \emph{unknown} constant, $\varnothing$, and the remaining half is distributed uniformly across all constants that can fill the role.

% Belief state updates
The belief states for all roles are initialized to uniform probabilities over constants.\footnote{Half the mass of non-{\it action} roles is initialized on the $\varnothing$ constant, a prior indicating that the role is not relevant for the not-yet-specified action.}
We denote the beliefs from a single utterance, $x$, as $\mathcal{B}_x$, itself a mapping from semantic roles to the distribution over constants that can fill them.
The agent's belief is updated with
\begin{equation}
	\label{eq:phm_belief_update}
	\mathcal{B}(r, a)\leftarrow (1 - \rho)\mathcal{B}(r, a) + \rho\mathcal{B}_x(r, a),
\end{equation}
for every semantic role $r$ and every constant $a$. % (Figure~\ref{fig:phm_belief_update_example}).
The parameter $\rho\in[0,1]$ controls how much to weight the new information against the current belief.\footnote{We set $\rho=0.5$ for clarification updates.}

% Sampling, dialog policy
After a belief update from a user response, the highest-probability constants for every semantic role in the current belief state $\mathcal{B}$ are used to select a question that the agent expects will maximize information gain.
Table~\ref{tab:phm_dialog_policy_examples} gives some examples of the policy $\pi$.

% Confirmation updates.
For updates based on confirmation question responses, the confirmed $\mathcal{B}_x$ constant(s) receive the whole probability mass for their roles (i.e., $\rho=1$).
% The update in Equation~\ref{eq:phm_belief_update} is performed with $\rho=1$, such that $\mathcal{B}$ reflects the confirmation.
If a user denies a confirmation, $\mathcal{B}_x$ is constructed with the constants in the denied question given zero probability for their roles, and other constants given uniform probability (so Equation~\ref{eq:phm_belief_update} reduces the belief only for denied constants).
A conversation concludes when the user has confirmed every semantic role.
% Multiple roles can be confirmed at once, for example when asking the user to confirm the whole command.

% \begin{figure}
% \centering
% \begin{subfigure}{0.23\textwidth}
%   \centering
%   \includegraphics[width=1.\linewidth]{figures/interface_oal_yn_ex.png}
% \end{subfigure}
% \begin{subfigure}{0.23\textwidth}
%   \centering
%   \includegraphics[width=\linewidth]{figures/interface_oal_pos_ex.png}
% \end{subfigure}
% \caption{The agent uses opportunistic active learning to select questions about whether a concept applies.
% }
%   \label{fig:web_interface_oal}
% \end{figure}

%% file: 0304_agent_learning.tex
The agent improves its semantic parser by inducing training data over finished conversations.
Perceptual concept models are augmented on the fly from questions asked to a user, and are then aggregated across users in batch.

\paragraph{Semantic Parser Learning From Conversations}
The agent identifies utterance-denotation pairs in conversations by pairing the user's initial command with the final confirmed action, and answers to questions about each role with the confirmed role (e.g., \emph{robert's office} as the \emph{goal} location $r_1$), similar to prior work~\cite{thomason:ijcai15}.
Going beyond prior work, the agent then finds the latent parse for the pair: a beam of parses is created for the utterance, and these are grounded to discover those that match the target denotation.
% The agent selects from among matches the one parse with highest joint confidence between parsing and grounding (i.e., the parse derivable from the language text that also grounds to the known, correct confirmation).
The agent then retrains its parser given these likely, latent utterance-semantic parse pairs as additional, weakly-supervised examples of how natural language maps to semantic slots in the domain.

% Synonymy sub-dialogs.
% Opportunistic active learning-based perception sub-dialogs.
\paragraph{Opportunistic Active Learning}
Some unseen words are perceptual concepts.
If one of the neighboring words of unknown word $x_i$ is associated with a semantic form involving a perceptual concept, the agent asks: \emph{I haven't heard the word `$x_i$' before. Does it refer to properties of things, like a color, shape, or weight?}
If confirmed, the agent ranks the nearest neighbors of $x_i$ by distance and sequentially asks the user whether the next nearest neighbor $t_p$ is a synonym of $x_i$.
If so, new lexical entries are created to allow $x_i$ to function like $t_p$, including sharing an underlying concept model (e.g., in our experiments, \emph{tall} was identified as a synonym of the already-known word \emph{long}).
Otherwise, a new concept model is created for $x_i$ (e.g., in our experiments, the concept \emph{red}).

We introduce opportunistic active learning questions~\cite{thomason:corl17} as a sub-dialog, in which the agent can query about training objects \emph{local} to the human and the robot (Figure \ref{fig:segbot_demo}).
This facilitates on the fly acquisition of new concepts, because the agent can ask the user about nearby objects, then apply the learned concept to \emph{remote} test objects (Section~\ref{ssec:experiment_robot}).

%% file: 04_experiments.tex
% Overview.
We hypothesize that the learning capabilities of our agent will improve its language understanding and usability.
We also hypothesize that the agent trained in a simplified world simulation on Mechanical Turk can be deployed on a physical robot, and can learn non-visual concepts (e.g., {\it rattling}) on the fly that could not be acquired in simulation.

\subsection{Experiment Design}
\label{ssec:experiment_design}
\input{0401_experiment_design}

\begin{figure}[t]
\centering
  \includegraphics[width=1.\linewidth]{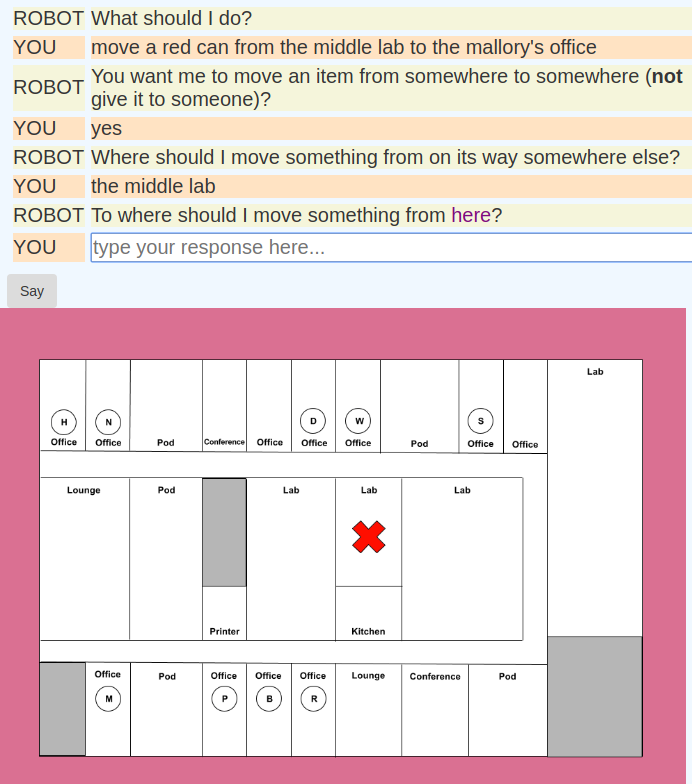}
\caption{The agent asks questions to clarify the command through dialog.
Each clarification is used to induce weakly-supervised training examples for the agent's semantic parser.}
  \label{fig:phm_web_interface_move}
\end{figure}

\subsection{Mechanical Turk Evaluation}
\label{ssec:experiment_mturk}
\input{0402_experiment_mturk}

\subsection{Physical Robot Implementation}
\label{ssec:experiment_robot}
\input{0403_experiment_robot}

%% file: 0401_experiment_design.tex
The agent (and corresponding robot) can perform three high-level tasks: \emph{navigation} (the agent goes to a location), \emph{delivery} (the agent takes an object to a person), and \emph{relocation} (the agent takes an object from a source location to a goal location).
We denote 8 (randomly selected) of the 32 objects explored in prior work~\cite{sinapov:ijcai16} as test objects and the remaining 24 as training objects available for active learning queries.
We randomly split the set of possible task instantiations (by room, person, and object arguments) into initialization (10\%), train (70\%), and test sets (20\%).

%Discuss initialization phase, bootstrapping parser lexicon, ontology, and init paired examples.
\paragraph{Initialization Phase}
Sixteen users (graduate students) were shown one of each type of task (from the initialization set) and gave two high-level natural language commands for each (initial and rephrasing).
We used a subset of these utterances\footnote{Commands that would introduce rare predicates were dropped.} as a scaffold on which to build a seed language-understanding pipeline: an initial lexicon and a set of 44 utterance-semantic parse pairs, $D_0$.\footnote{An experimenter performed the annotations to create these resources in about two hours.}

% Motivate and present the three training folds, then discuss retraining procedure used between each, then final testing fold and how it differs.
\paragraph{Training Procedure}
The initial pipeline is used by a baseline agent $\mathcal{A}_1$; we denote its parser $\mathcal{P}_1$ trained on $D_0$, and denote untrained concept models for several predicates $P_{c,1}$.
That is, the initial lexicon contains several concept words (like \emph{yellow}), but no labels between objects and these concepts.
All learning for the parsing and perception modules arises from human-agent conversations.

% Three training folds to get wider range of lexical use and new primer training objects
We divide the training procedure into three phases, each associated with 8 different objects from the active training set of 24.
The perceptual concept models are retrained on the fly during conversations as questions are asked (e.g., as in Figure~\ref{fig:rattling_highlight}).
The parsing model is retrained between phases.
Each phase $i$ is carried out by agent $\mathcal{A}_i$, and training on all phase conversations yields agent $\mathcal{A}_{i+1}$ using concept models $\mathcal{P}_{i+1}$ and parser $P_{c,i+1}$.
In each phase of training, and when evaluating agents in different conditions, we recruit 150 Mechanical Turk workers with a payout of \$1 per HIT.

% Metrics extracted from conversations (clarification questions, semantic role f1, surveys)
\paragraph{Testing and Performance Metrics}
We test agent $\mathcal{A}_3$ with parser $\mathcal{P}_3$ and perception models $P_{c,3}$ against unseen tasks,\footnote{Empirically, parser $\mathcal{P}_4$ overfits the training data, so we evaluate with $\mathcal{P}_3$. For $\mathcal{A}_4^*$, this is not a concern since the initial parser parameters $\mathcal{P}_1^*$ are used.} and denote it \emph{Trained (Parsing+Perception)}.
We also test an ablation agent, $\mathcal{A}_4^*$, with parser $\mathcal{P}_1^*$ and perception models $P_{c,4}$ (trained perception models with an initial, baseline parser with parsing rules only added for new concept model words), and denote it \emph{Trained (Perception)}.
These agents are compared against the baseline agent $\mathcal{A}_1$, denoted \emph{Initial (Seed)}.

We measure the number of clarification questions asked during the dialog to accomplish the task correctly.
This metric should decrease as the agent refines its parsing and perception modules, needing to ask fewer questions about the unseen locations and objects in the test tasks.
We also consider users' answers to survey questions about usability.
Each question was answered on a 7-point Likert scale: from \emph{Strongly Disagree} (1) to \emph{Strongly Agree} (7).

%% file: 0402_experiment_mturk.tex
We prompt users with instructions like: \emph{Give the robot a command to solve this problem: The robot should be at the X marked on the green map}, with a green-highlighted map marking the target.
Users are instructed to command the robot to perform a \textit{navigation}, \textit{delivery}, and \textit{relocation} task in that order.
The simple, simulated environment in which the instructions were grounded reflects a physical office space, allowing us to transfer the learned agent into an embodied robot (Section~\ref{ssec:experiment_robot}).
Users type answers to agent questions or select them from menus (Figure~\ref{fig:phm_web_interface_move}).
For \emph{delivery} and \emph{relocation}, target objects are given as pictures.
Pictures are also shown alongside concept questions like \emph{Would you use the word `rattling' when describing this object?}

% Boring details about dialog length soft caps, time limits, vetting for res ults (no-repeat, no-walk), and table showing, per condition: number of users who submitted the HIT, number of users validated, number of users vetted for inclusion in results; per task: correct and used for retraining; using word embeddings
% \paragraph{Human Intelligence Tasks (HITs)}
% Table~\ref{tab:phm_mturk_breakdown} gives the numbers of workers who engaged with our HITs.

% The low number of workers that complete the tasks given that they submitted the HIT at all suggests how difficult the HIT is, and will inform future experimental design.

% \begin{table*}
% \centering
% \begin{tabular}[ht]{|l|rrr|rrr|}
% 	\hline
% 	\bf Condition & \multicolumn{6}{c|}{\bf Number of Workers} \\ \hline
% 	& Submitted & Completed & Vetted & Nav. & Del. & Rel. \\
% 	& HIT & Tasks & & Correct & Correct & Correct \\ \hline \hline
% 	Train ($A_1, A_2, A_3$) & 297 & 162 & 113 & 36 & 44 & 18 \\ \hline
% 	Untrained ($A_1$) & 150 & 67 & 44 & 17 & 22 & 10 \\
% 	Test$^* (A_4^*)$ & 148 & 83 & 50 & 20 & 29 & 10 \\
% 	Test ($A_4$) & 143 & 79 & 42 & 16 & 23 & 10 \\
% 	\hline
% \end{tabular}
% \caption{We count only workers that {\bf submitted} the HIT with the correct code.
% Workers that {\bf completed} all tasks and the survey finished the HIT entirely.
% {\bf Vetted} workers' data was kept for evaluation.}
% \label{tab:phm_mturk_breakdown}
% \end{table*}

% Quantitative results (dialog length, f1)
\begin{table}[t]
    \centering
    \begin{tabular}{l r r r}
        \multirow{2}{*}{Agent} & \multicolumn{3}{c}{Clarification Questions $\downarrow$} \\
        & \multicolumn{1}{c}{Navigation $(p)$} & \multicolumn{1}{c}{Delivery $(p)$} & \multicolumn{1}{c}{Relocation $(p)$} \\
        \toprule
        In & $3.02\pm6.48$\phantom{$(.00)$} & $6.81\pm8.69$\phantom{$(.00)$} & $22.3\pm9.15$\phantom{$(.00)$} \\
        Tr$^*$ & $4.05\pm8.81 (.46)$ & $8.16\pm13.8 (.53)$ & $23.5\pm6.07 (.67)$ \\
        Tr & $1.35\pm4.44 (.11)$ & $7.50\pm9.93 (.72)$ & $19.6\pm7.89 (.47)$ \\
        \bottomrule
    \end{tabular}
    \caption{The average number of clarification questions agents asked among dialogs that reached the correct task.
    Also given are the $p$-values of a Welch's $t$-test between the \textbf{Tr}\emph{ained}$^*$ (\emph{Perception}) and \textbf{Tr}\emph{ained} (\emph{Parsing+Perception}) model ratings against the \textbf{In}\emph{itial} model ratings.
    }
    \label{tab:phm_quant_results}
\end{table}
% TODO: discuss
    % Positive: P+P achieves lowest clarification qs for navigation and relocation.
    % Negative: P-only sees small bump in turns across board; P+P hurts delivery over initial maybe?
    
Table \ref{tab:phm_quant_results} gives measures of the agents' performance in terms of the number of clarification questions asked before reaching the correct task specification to perform.
For both \emph{navigation} and \emph{relocation}, there is a slight decrease in the number of questions between the \emph{Initial} agent and the \emph{Trained (Parsing+Perception)} agent.
The \emph{Trained (Perception)} agent which only retrains and adds new concept models from conversation history sees slightly worse performance across tasks, possibly due to a larger lexicon of adjectives and nouns (e.g., \emph{can} as a descriptive noun now polysemous with \emph{can} as a verb---\emph{can you...}) without corresponding parsing updates.
None of these differences are statistically significant, possibly because comparatively few users completed tasks correctly, necessary to use this metric.\footnote{Across agents, an average of 42\%, 39\%, and 9.5\% workers completed \textit{navigation}, \textit{delivery}, and \textit{relocation} tasks correctly, respectively. A necessary step in future studies is to improve worker success rates, possibly through easier interfaces, faster HITs, and higher payouts.}

% Qualitative results (user surveys)
\begin{table}[t]
    \centering
    \begin{tabular}{l r r r}
        \multirow{2}{*}{Agent} & \multicolumn{3}{c}{Usability Survey (Likert 1-7) $\uparrow$} \\
        & \multicolumn{1}{c}{Navigation $(p)$} & \multicolumn{1}{c}{Delivery $(p)$} & \multicolumn{1}{c}{Relocation $(p)$} \\
        \toprule
        In & $3.09\pm2.04$\phantom{$(.00)$} & $3.20\pm2.12$\phantom{$(.00)$} & $3.37\pm2.17$\phantom{$(.00)$} \\
        Tr$^*$ & $3.51\pm2.05 (.09)$ & $3.60\pm2.09 (.12)$ & $3.60\pm2.08 (.37)$ \\
        Tr & $\pmb{3.76}\pm2.07 (.01)$ & $\pmb{3.87}\pm2.10 (.01)$ & $\pmb{3.93}\pm2.16 (.04)$ \\
        \bottomrule
    \end{tabular}
    \caption{The average Likert rating given on usability survey prompts for each task across the agents.
    {\bf Bold} indicates an average \textbf{Tr}\emph{ained}$^*$ (\emph{Perception}) and \textbf{Tr}\emph{ained} (\emph{Parsing+Perception}) model ratings significantly higher than the \textbf{In}\emph{itial} model ($p<0.05$) under a Welch's $t$-test.
    }
    \label{tab:phm_qual_results}
\end{table}

Table \ref{tab:phm_qual_results} gives measures of the agents' performance in terms of qualitative survey prompt responses from workers.
Prompts were: {\it I  would  use  a  robot  like  this  to  help navigate a new building}, {\it I would use a robot like this to get items for myself or others}, and {\it I would use a robot like this to move items from place to place}.
Across tasks, the \emph{Trained (Parsing+Perception)} agent novel to this work is rated as more usable than both the \emph{Initial} agent and the \emph{Trained (Perception)} agent that updated only its concept models from training conversations.

\begin{figure}
\centering
\begin{tabular}[t!]{cccc}
	\multicolumn{4}{c}{\bf Learned Concept Model for \emph{can}} \\
	\includegraphics[width=0.15\linewidth]{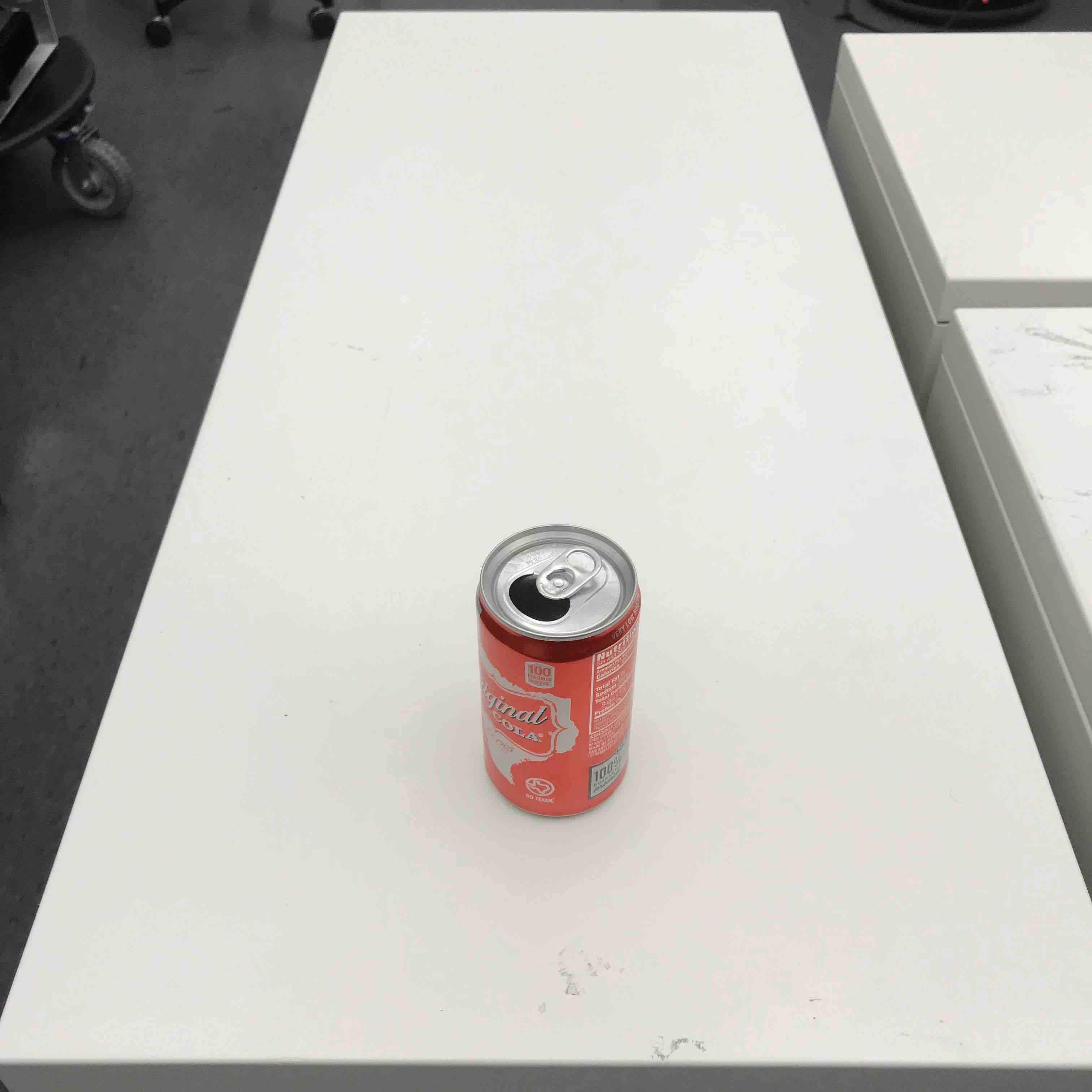} &
	\includegraphics[width=0.15\linewidth]{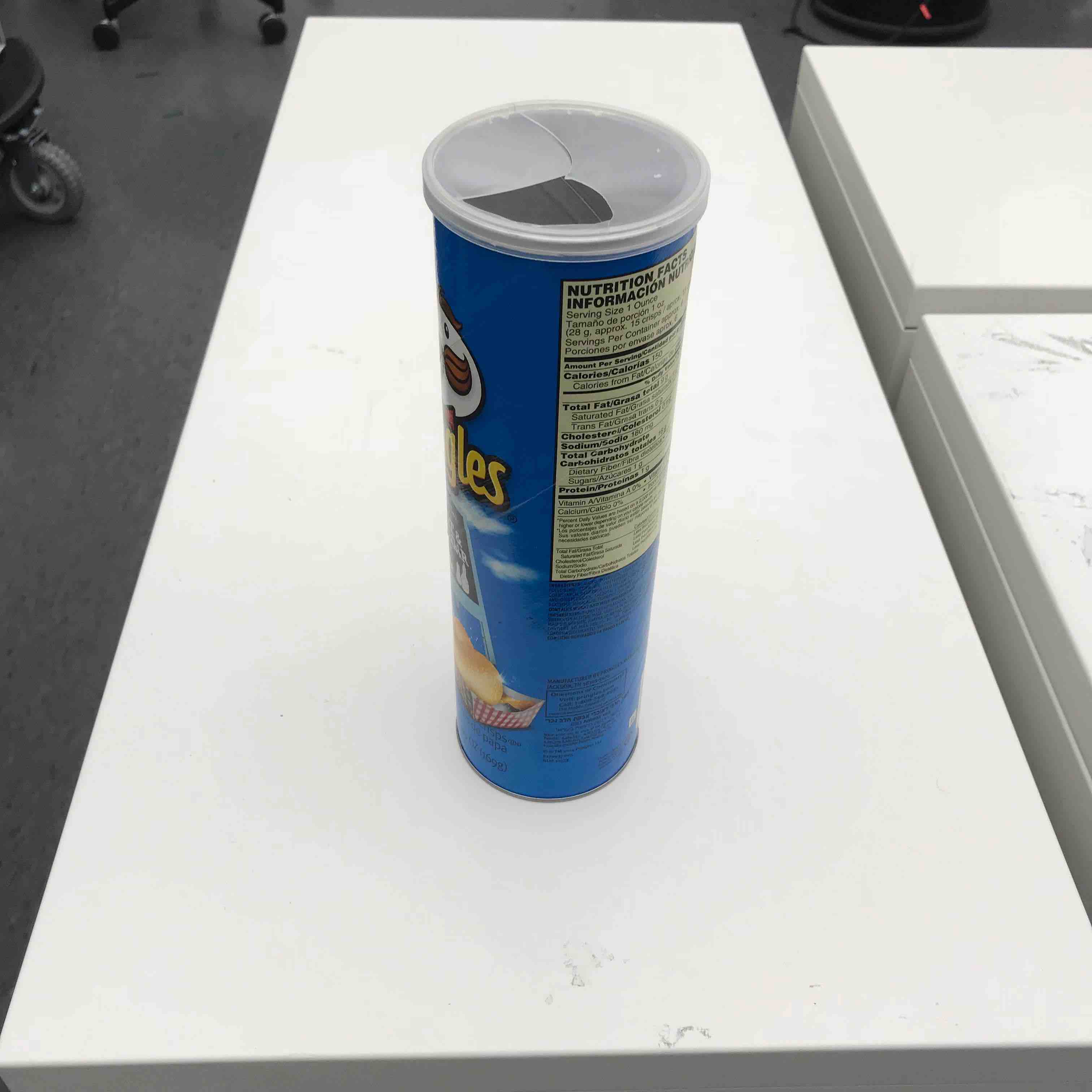} &
	\includegraphics[width=0.15\linewidth]{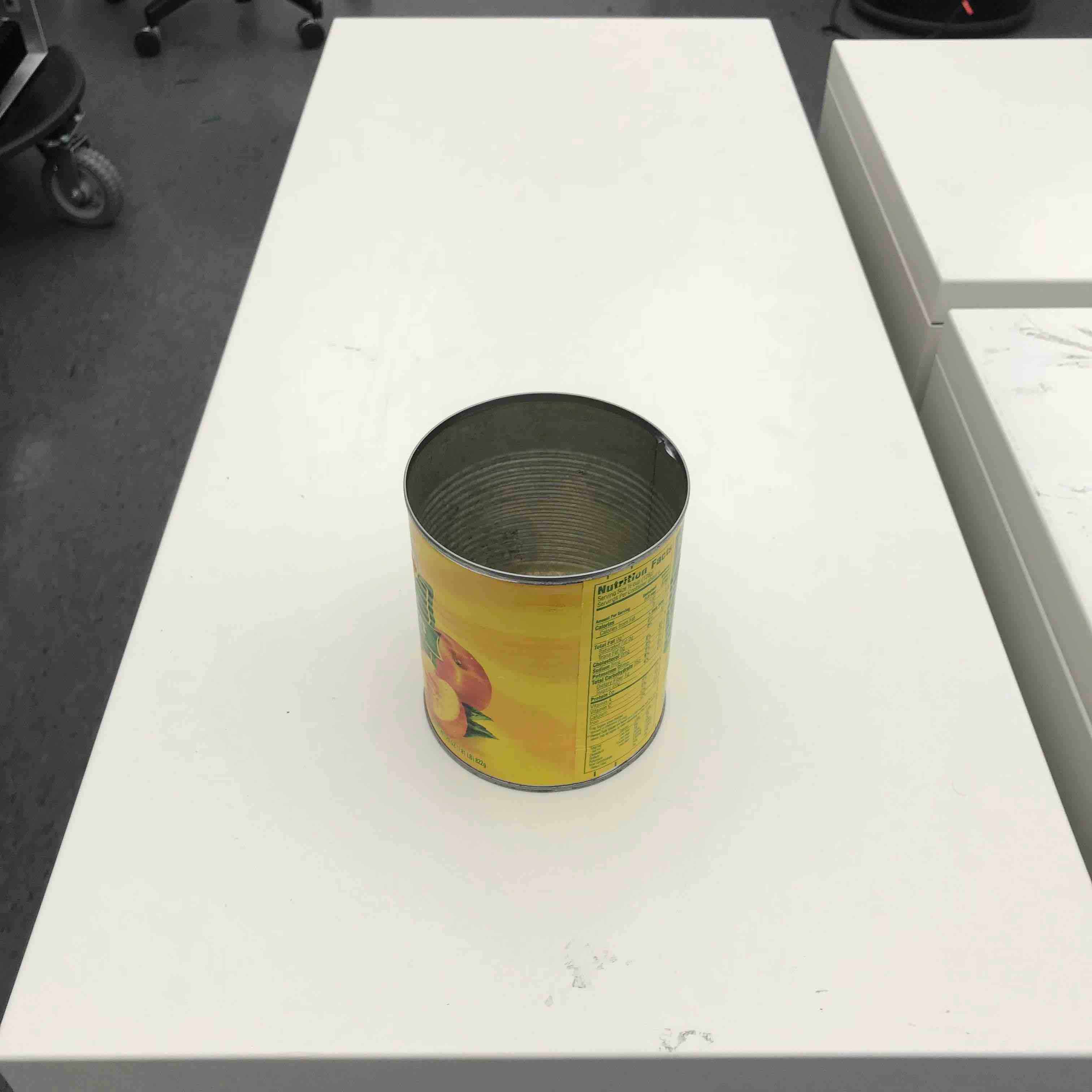} &
	\includegraphics[width=0.15\linewidth]{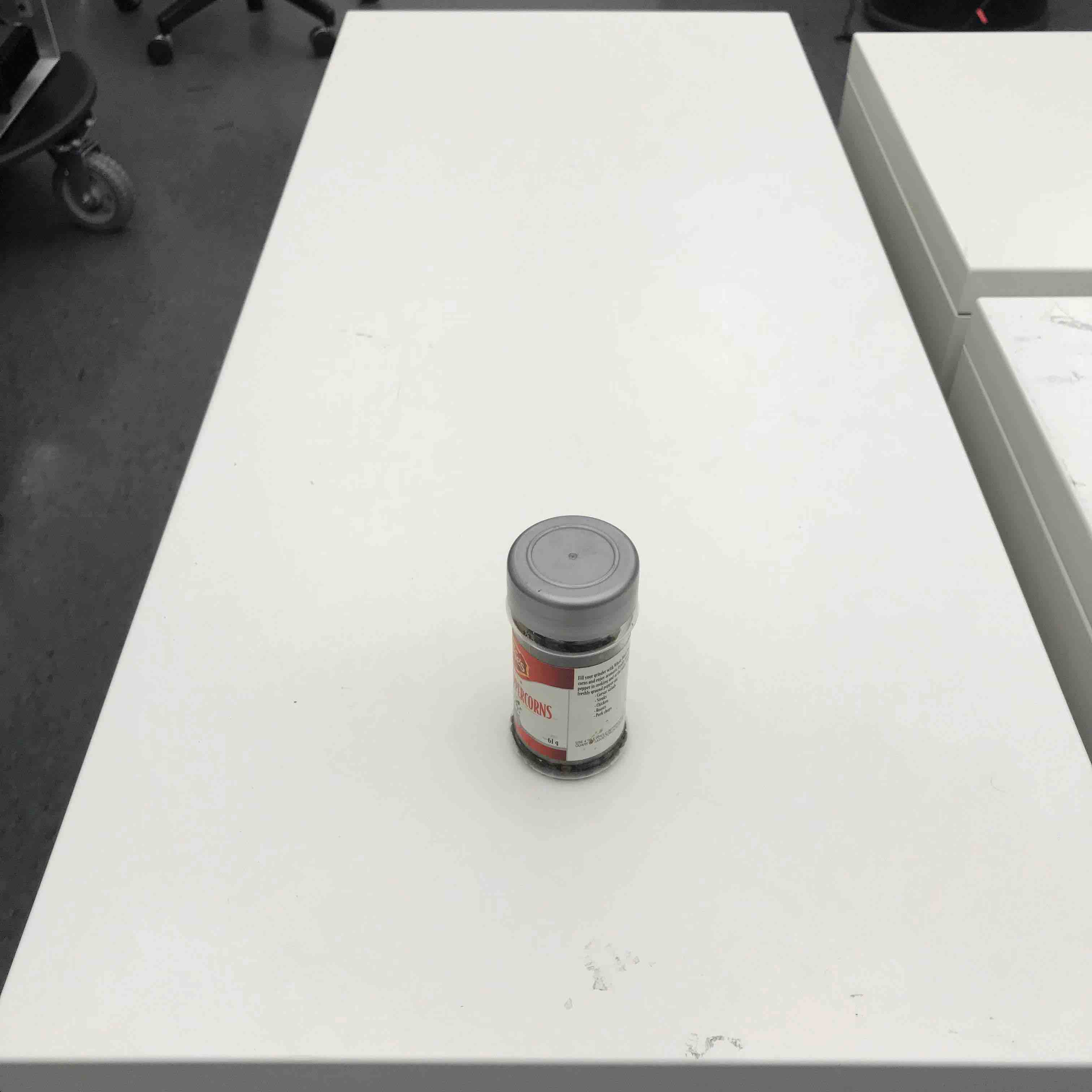} \\
	0.32 & 0.22 & 0.2 & 0.13 \\
	\includegraphics[width=0.15\linewidth]{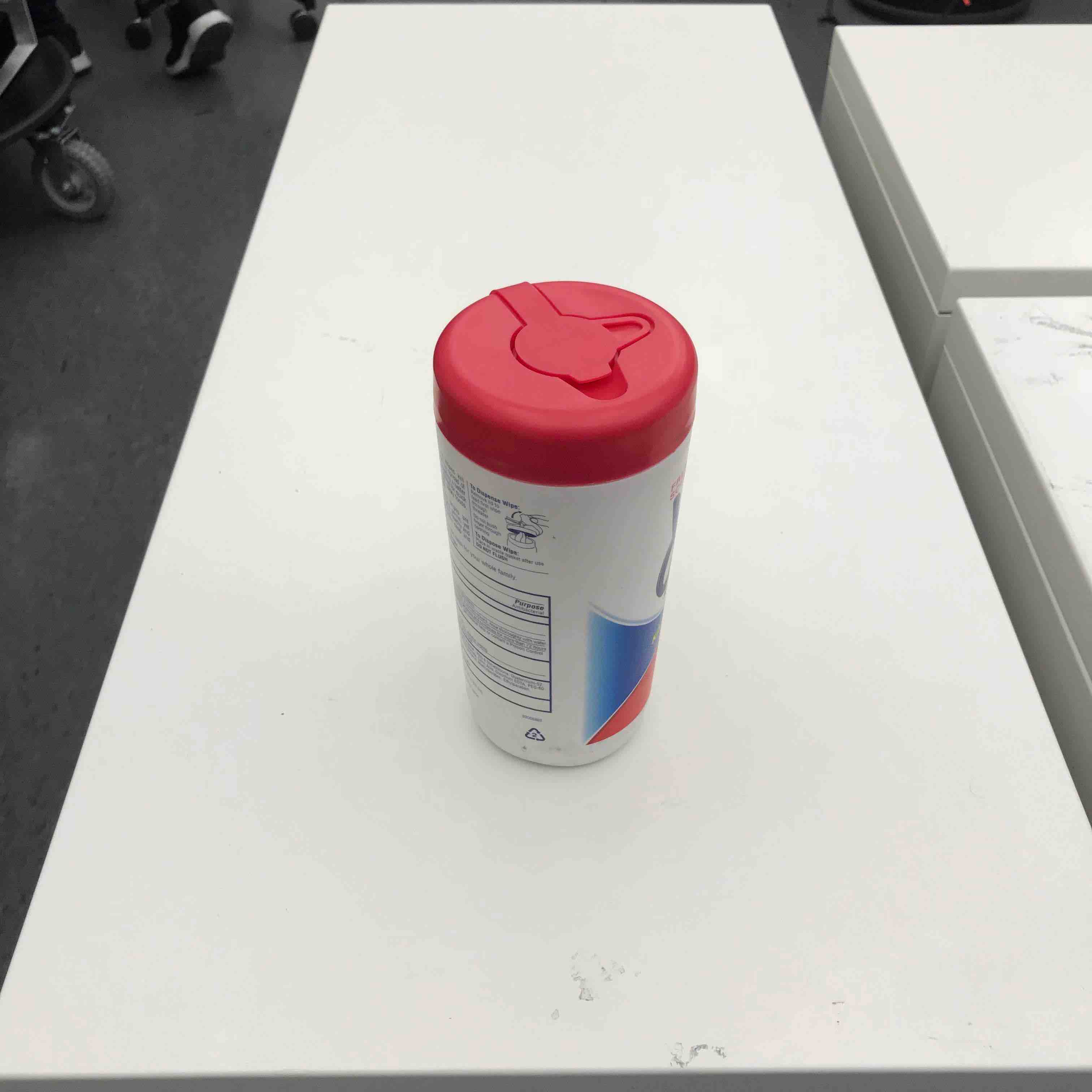} &
	\includegraphics[width=0.15\linewidth]{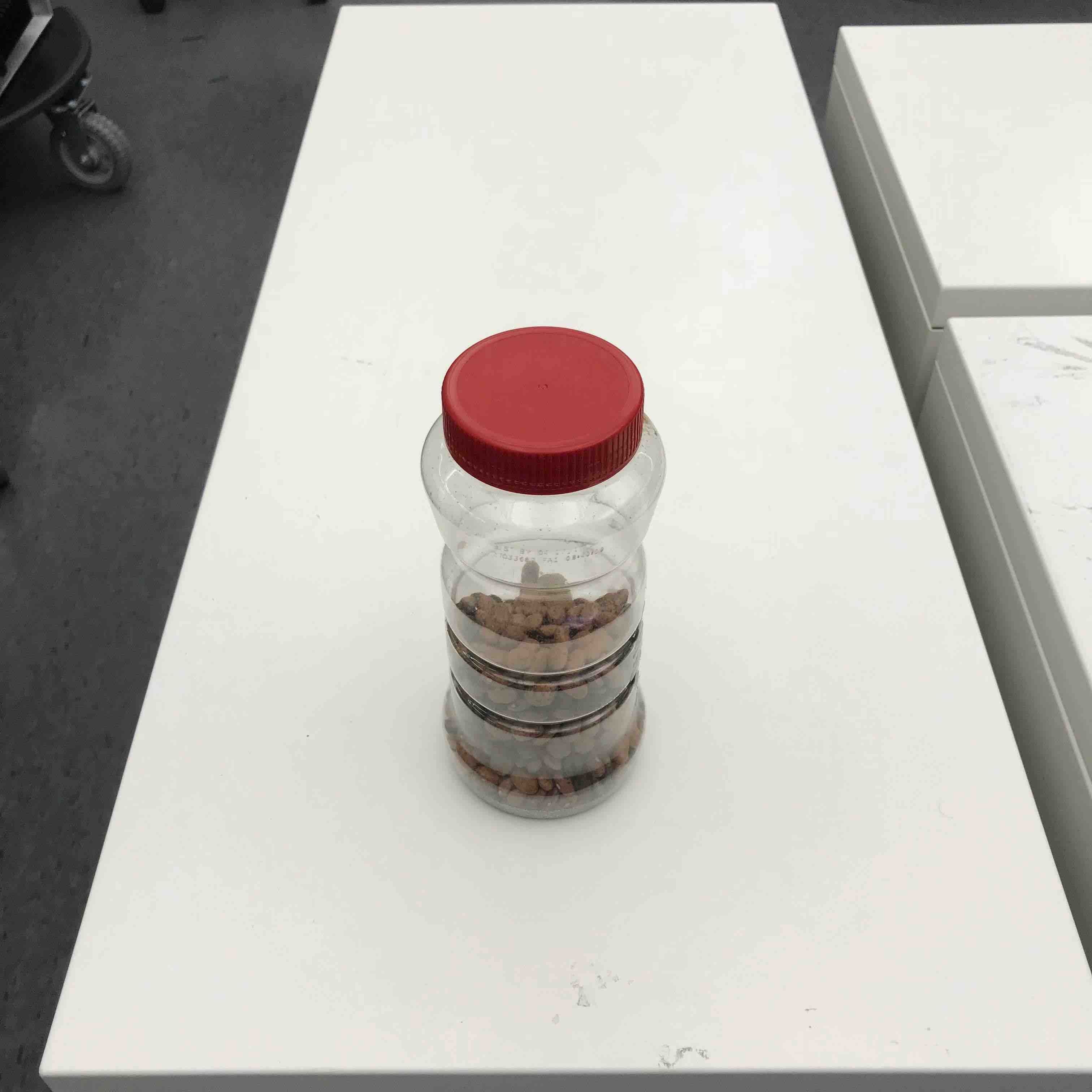} &
	\includegraphics[width=0.15\linewidth]{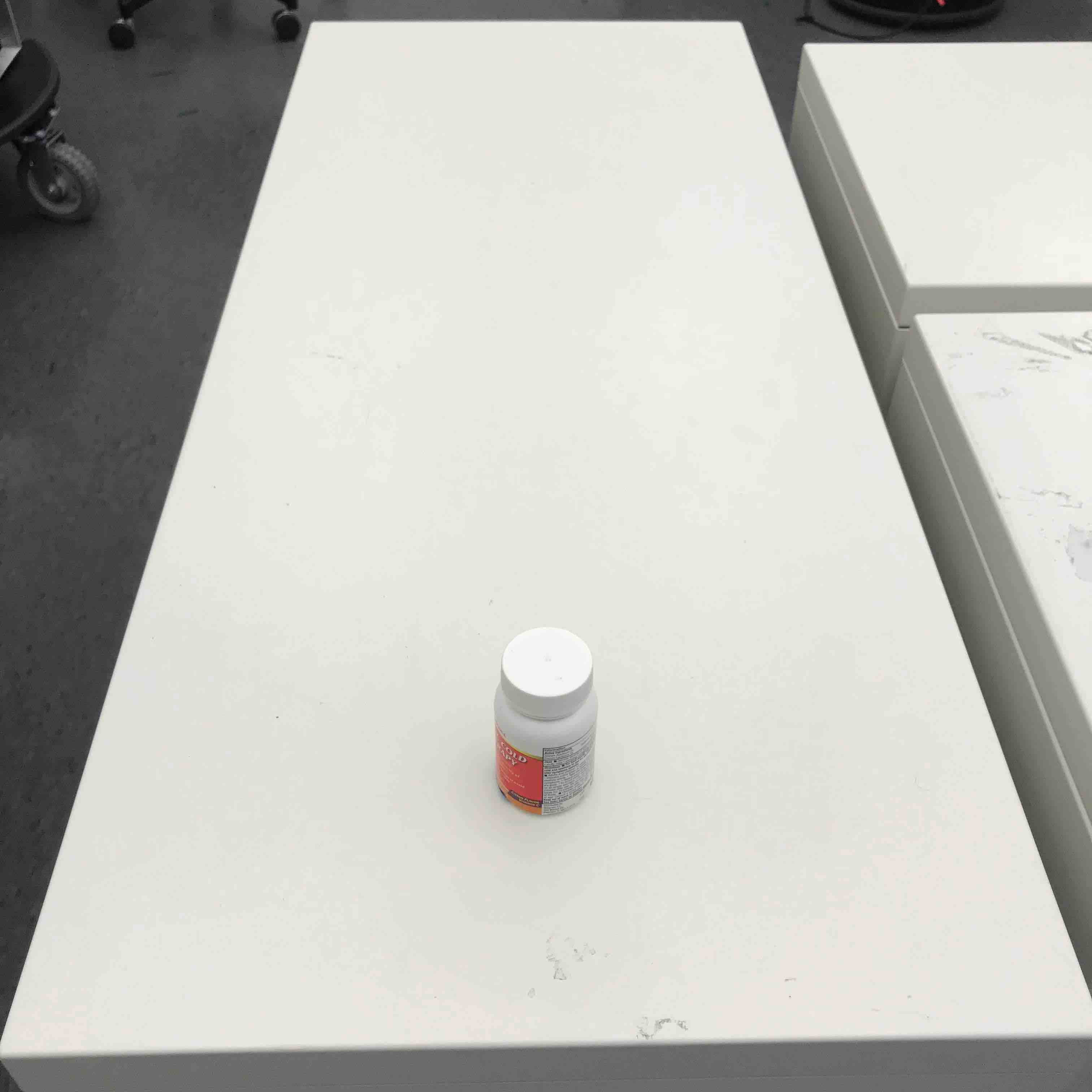} &
	\includegraphics[width=0.15\linewidth]{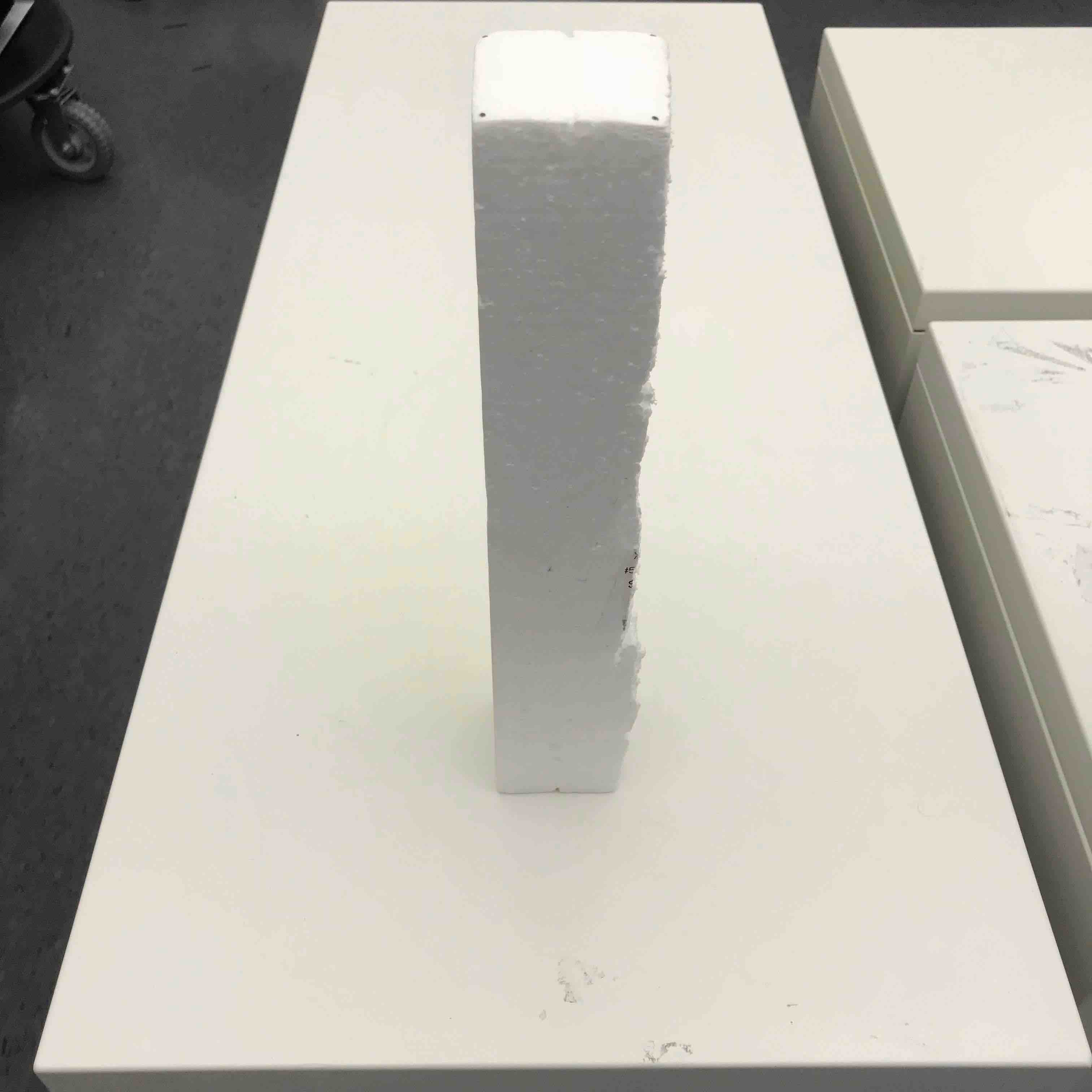} \\
	0.07 & 0.03 & 0.03 & 0 \\
\end{tabular}
\caption{Confidence distribution for the \emph{can} concept model on the unseen test set objects after training.}
\label{fig:concept_model_can}
\end{figure}

The agent acquired new perceptual concept models (25 in total), and synonym words for existing concepts during training.
% Mechanical Turk workers are sometimes inattentive in the long HIT, so the agent's learned concept models are imperfect.
Figure~\ref{fig:concept_model_can} shows the learned model for \emph{can} on unseen test objects.
The agent's ordering of test objects' \emph{can}-ness qualitatively matches intuition.

%% file: 0403_experiment_robot.tex
\begin{figure*}
\centering
\begin{subfigure}{1.\textwidth}
  \centering
  \begin{tabular}{cc}
    \begin{tabular}{ll}
      \\
      \textcolor{Red}{{\bf R}}:& \textcolor{Red}{What should I do?} \\
      \textcolor{Blue}{{\bf H}}:& \textcolor{Blue}{Move a rattling container from the} \\
      & \textcolor{Blue}{lounge by the conference room to} \\
      & \textcolor{Blue}{Bob's office.} \\
      \textcolor{Red}{{\bf R}}:& \textcolor{Red}{Does rattling refer to a property of} \\
      & \textcolor{Red}{an object?} \\
      \textcolor{Blue}{{\bf H}}:& \textcolor{Blue}{Yes.} \\
      \textcolor{Red}{{\bf R}}:& \textcolor{Red}{Does rattling mean the same thing} \\
      & \textcolor{Red}{as empty?} \\
      \textcolor{Blue}{{\bf H}}:& \textcolor{Blue}{No.} \\
      \textcolor{Red}{{\bf R}}:& \textcolor{Red}{I am still learning the meanings of} \\
      & \textcolor{Red}{some words.} \\
      \textcolor{Red}{{\bf R}}:& \textcolor{Red}{Show me an object you could use} \\
      & \textcolor{Red}{the word rattling when describing,} \\
      & \textcolor{Red}{or say ``none of them.''} \\
      \textcolor{Blue}{{\bf H}}:& \textcolor{Blue}{Watch.} \\
      \textcolor{Red}{{\bf R}}:& \textcolor{Red}{I am getting the objects on the table} \\
      & \textcolor{Red}{into focus$\dots$ Okay, I see them.} \\
      \textcolor{Blue}{{\bf H}}:& \textcolor{Blue}{This one. \emph{The human picks up a}} \\
      & \textcolor{Blue}{\emph{small, full pill jar.}} \\
    \end{tabular} &
    \begin{tabular}{ll}
    \\
      \textcolor{Red}{{\bf R}}:& \textcolor{Red}{I see.} \\
      \textcolor{Red}{{\bf R}}:& \textcolor{Red}{Show me an object you could not} \\
      & \textcolor{Red}{use the word rattling when} \\
      & \textcolor{Red}{describing, or say ``all of them.''} \\
      \textcolor{Blue}{{\bf H}}:& \textcolor{Blue}{Watch. This one. \emph{The human picks}} \\
      & \textcolor{Blue}{\emph{up a red water bottle.}} \\
      \textcolor{Red}{{\bf R}}:& \textcolor{Red}{I see.} \\
      \textcolor{Red}{{\bf R}}:& \textcolor{Red}{Show me an object you could use} \\
      & \textcolor{Red}{the word rattling when describing,} \\
      & \textcolor{Red}{or say ``none of them.''} \\
      \textcolor{Blue}{{\bf H}}:& \textcolor{Blue}{Watch. This one. \emph{The human picks}} \\
      & \textcolor{Blue}{\emph{up a container half-full of beans.}} \\
      \textcolor{Red}{{\bf R}}:& \textcolor{Red}{I see.} \\
      \textcolor{Red}{{\bf R}}:& \textcolor{Red}{Thanks. Now back to business.} \\
      \textcolor{Red}{{\bf R}}:& \textcolor{Red}{You want me to move an item} \\
      & \textcolor{Red}{from 3.514 to somewhere?} \\
      \textcolor{Blue}{{\bf H}}:& \textcolor{Blue}{Yes.} \\
      \textcolor{Red}{{\bf R}}:& \textcolor{Red}{I should move something from} \\
      & \textcolor{Red}{3.514 to 3.510? [\emph{room numbers}]} \\
      \textcolor{Blue}{{\bf H}}:& \textcolor{Blue}{Yes.}
    \end{tabular}
  \end{tabular}
  \label{fig:segbot_demo:conv}
\end{subfigure}
\begin{subfigure}{0.25\textwidth}
  \centering
  \includegraphics[width=\linewidth]{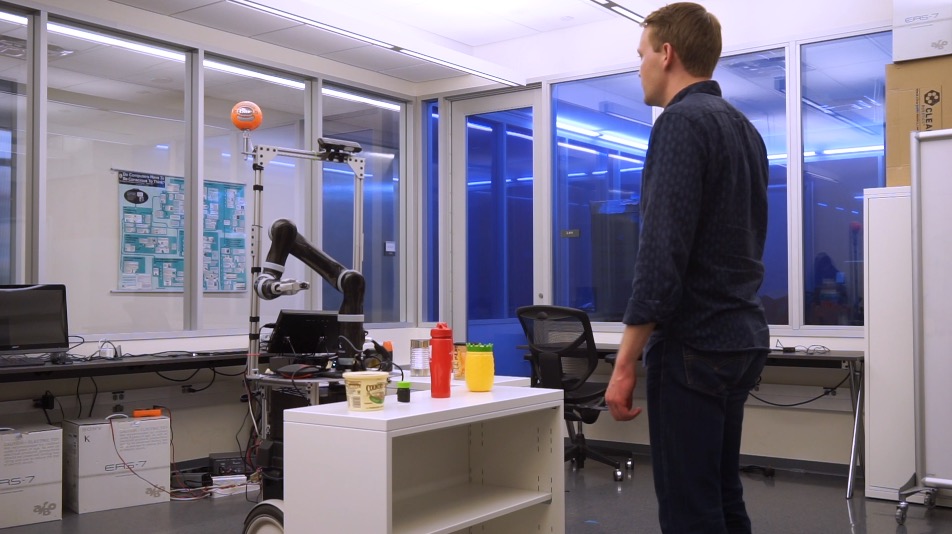}
  \caption{The human says \emph{move a rattling container}.}
  \label{fig:segbot_demo:1}
\end{subfigure}
\begin{subfigure}{0.25\textwidth}
  \centering
  \includegraphics[width=\linewidth]{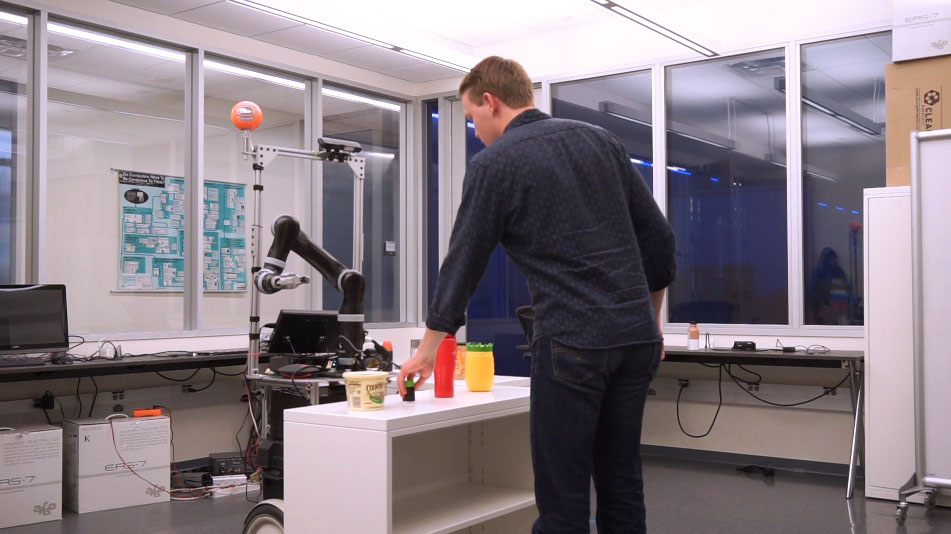}
  \caption{The robot asks questions about local items to learn \emph{rattling}.}
  \label{fig:segbot_demo:2}
\end{subfigure}
\begin{subfigure}{0.25\textwidth}
  \centering
  \includegraphics[width=\linewidth]{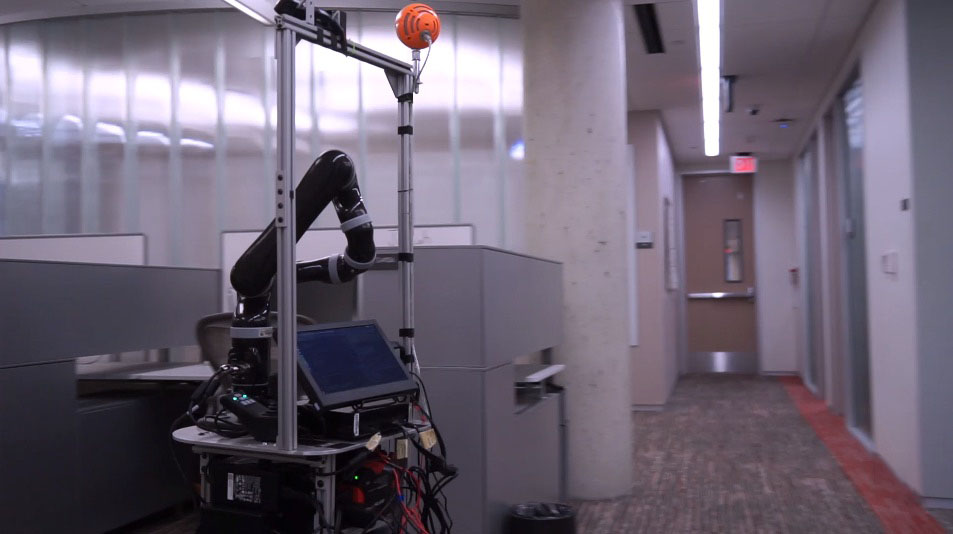}
  \caption{The robot moves to the specified location.}
  \label{fig:segbot_demo:3}
\end{subfigure}
\begin{subfigure}{0.25\textwidth}
  \centering
  \includegraphics[width=\linewidth]{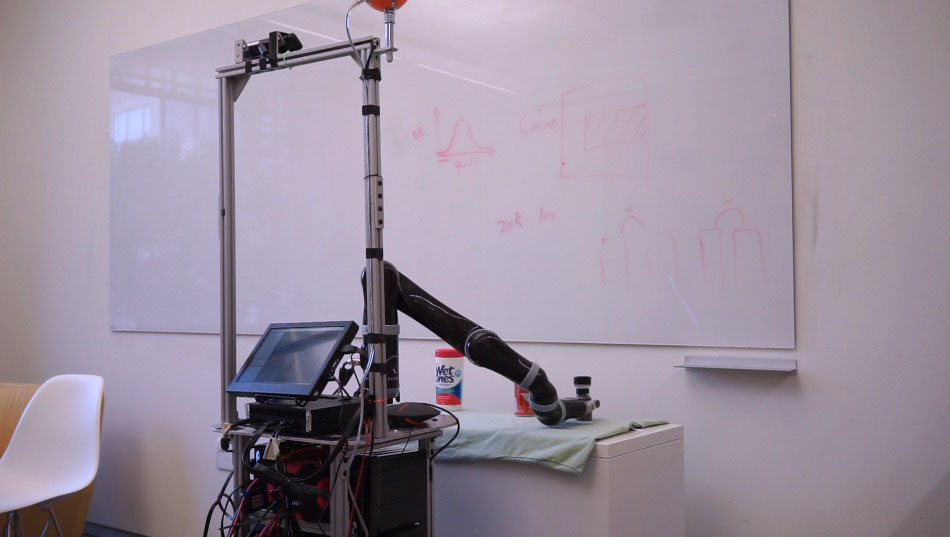}
  \caption{The robot infers and grasps a \emph{rattling container}.}
  \label{fig:segbot_demo:4}
\end{subfigure}
\begin{subfigure}{0.25\textwidth}
  \centering
  \includegraphics[width=\linewidth]{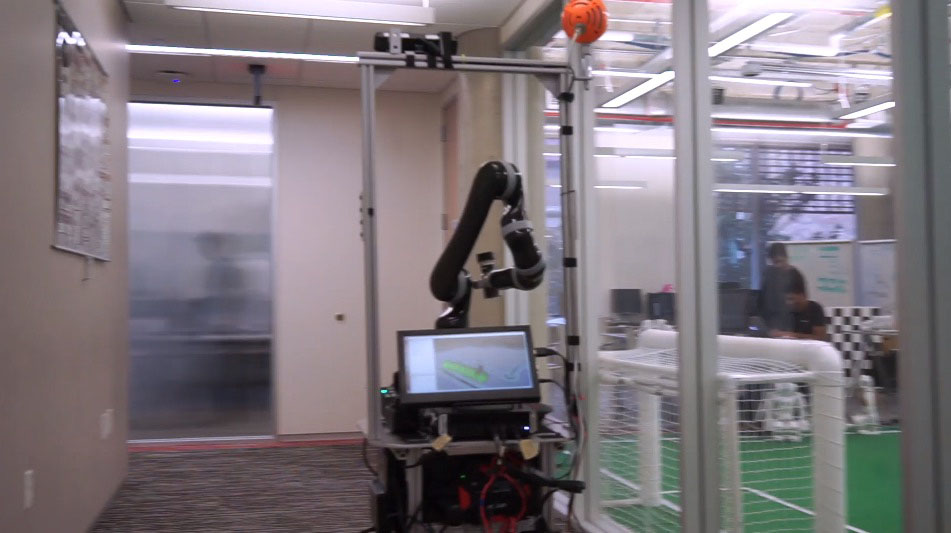}
  \caption{The robot navigates to the specified destination room.}
  \label{fig:segbot_demo:5}
\end{subfigure}
\begin{subfigure}{0.25\textwidth}
  \centering
  \includegraphics[width=\linewidth]{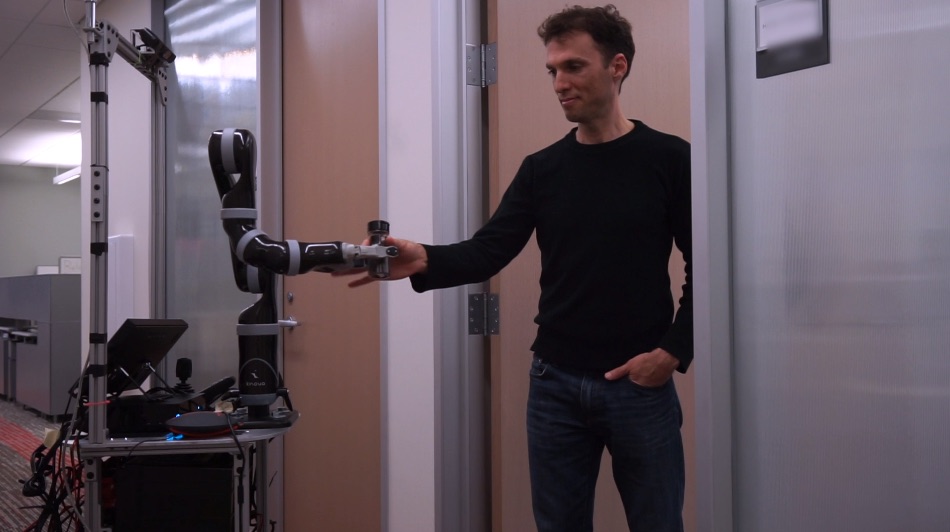}
  \caption{The robot hands over the item at the specified destination.}
  \label{fig:segbot_demo:6}
\end{subfigure}
\caption{The \emph{Trained (Parsing+Perception)} agent continues learning on the fly to achieve the specified goal.}
  \label{fig:segbot_demo}
\end{figure*}

The browser-interfaced, Mechanical Turk agent enabled us to collect controlled training data, but our end goal is a human-robot interface in a physically shared environment.
To establish that the agent and learning pipeline are robust and efficient enough to operate on real hardware in a live setting, we complement our Mechanical Turk evaluation with a demonstration of an embodied robot agent (Figure~\ref{fig:segbot_demo}).

We use the BWIBot~\cite{khandelwal:icaps14,khandelwal:ijrr17}, which can perceive and manipulate objects (Xtion ASUS Pro camera, Kinova MICO arm), navigate autonomously (Hokuyo lidar), record and transcribe human speech (Blue Snowball microphone, Google Speech API\footnote{\url{https://cloud.google.com/speech/}}), and verbalize audio responses (Festival Speech Synthesis System\footnote{\url{http://www.cstr.ed.ac.uk/projects/festival/}}).
Tabletop perception is implemented with RANSAC \cite{fischler:acm81} plane fitting and Euclidean clustering as provided by the Point Cloud Library~\cite{rusu:icra11}.

The agent is trained on Mechanical Turk conversations, transferring learned linguistic (e.g., \emph{lounge by the conference room}) and perceptual (e.g., object classes like \emph{can}) knowledge across platforms from simple simulation to real world application.
In a live human-robot dialog, an experimenter tells the agent to \emph{move a rattling container from the lounge by the conference room to bob's office}, requiring the agent to select correct rooms and to learn the new, audio-grounded word \emph{rattling} from the human user.\footnote{Demonstration video: \url{https://youtu.be/PbOfteZ_CJc?t=5}.}

%% file: 05_conclusion.tex
This paper proposes a robotic agent that leverages conversations with humans to expand small, hand-crafted language understanding resources both for translating natural language commands to abstract semantic forms and for grounding those abstractions to physical object properties.
We make several key assumptions, and promising areas of future work involve removing or weakening those assumptions.
In this work, the actions the robot can perform can be broken down into tuples of discrete semantic roles (e.g., \emph{patient, source}), but, in general, robot agents need to reason about more continuous action spaces, and to acquire new, previously unseen actions from conversations with humans~\cite{chai:ijcai18}.
When learning from conversations, we also assume the human user is cooperative and truthful, but detecting and dealing with combative users is necessary for real world deployment, and would improve learning quality from Mechanical Turk dialogs.
Making a closed world assumption, our agent has explored all available objects in the environment, but detecting and exploring objects on the fly using only task relevant behaviors~\cite{thomason:aaai18,IJCAI18-saeid} would remove this restriction.
Finally, dealing with complex adjective-noun dependencies (e.g., a \emph{fake gun} is \emph{fake} but is not a \emph{gun}) and graded adjectives (e.g., a \emph{heavy} mug weighs less than a \emph{light} suitcase) is necessary to move beyond simple, categorical object properties like \emph{can}.

We hope that our agent and learning strategies for an end-to-end dialog system with perceptual connections to the real world inspire further research on grounded human-robot dialog for command understanding.